\pdfoutput=1
\documentclass[11pt]{article}

\usepackage{arxiv}

\usepackage[T1]{fontenc}
\usepackage[utf8]{inputenc}
\usepackage{amsmath}
\usepackage{amssymb}
\usepackage{graphicx}
\usepackage{booktabs}
\usepackage{microtype}
\usepackage{siunitx}
\usepackage[numbers,sort&compress]{natbib}
\usepackage{hyperref}
\usepackage[capitalize]{cleveref}

\hypersetup{
  colorlinks=true,
  linkcolor=black,
  citecolor=black,
  urlcolor=blue,
  hypertexnames=false,
  pdftitle={CLIPPER: Replayable Shortlisted Optimization for Repeated Spatial Coverage Planning},
  pdfauthor={Julian Teusch, Joerg Philipp Mueller, Monika Sester}
}

\providecommand{\Description}[1]{}
\newenvironment{acks}{\section*{Acknowledgments}}{}
\renewcommand{\headeright}{CLIPPER Preprint}
\renewcommand{\shorttitle}{CLIPPER}
\renewcommand{\undertitle}{Author preprint. Accepted at ACM SIGSPATIAL 2026.}

\title{CLIPPER: Replayable Shortlisted Optimization for Repeated Spatial Coverage Planning}

\author{%
  Julian Teusch\textsuperscript{1}
  \quad
  J{\"o}rg Philipp M{\"u}ller\textsuperscript{1}
  \quad
  Monika Sester\textsuperscript{2}\\
  \textsuperscript{1}Institute of Computer Science, Clausthal University of Technology, Germany\\
  \textsuperscript{2}Institute of Cartography and Geoinformatics, Leibniz University Hannover, Germany\\
  \texttt{julian.teusch@tu-clausthal.de},
  \texttt{joerg.mueller@tu-clausthal.de}\\
  \texttt{monika.sester@ikg.uni-hannover.de}
}

\date{August 2026\\[0.2em]\small\href{https://doi.org/10.1145/3841645.3843432}{doi:10.1145/3841645.3843432}}

\begin{document}

\maketitle

\begin{abstract}
Operational requirements developed with the City of Braunschweig frame municipal micromobility
planning under geofenced exclusions, mandatory retained sites, spacing rules, and area-level caps.
Each policy edit requires a new feasible plan; full-set greedy takes tens of seconds per alternative
at city scale. We present \textbf{CLIPPER} (Constraint-exact Low-latency Iterative Planning with
Pooled Evaluation and Replay). It forms bounded
candidate pools but recomputes exact current gains and checks every active constraint before
selection. Coverage from each candidate alone sets the initial order. Offline full-set scans measure
gains omitted by the pool; online, a conservative bound triggers expansion or audit. CLIPPER-F gives
each proposal group the same number of candidate slots. Across Braunschweig, Munich, and Berlin, its
mean coverage over complete chains stays within \(0.245\) percentage points of full-set greedy under
the same policy, with \(13.6\text{--}28.9\times\) lower mean rollout time. CLIPPER-A instead
distributes one shared candidate budget across the groups. Under its coverage-prioritized policy, it uses
\(9\text{--}15\%\) of full-set greedy's rollout time under the same policy, with mean gaps of
\(1.82\) percentage points in Braunschweig, \(0.12\) in Munich, and \(0.27\) in Berlin.
Together, CLIPPER enables rapid, replayable comparison of recorded city-scale planning states while
enforcing every encoded model constraint.
\end{abstract}

\keywords{spatial decision support, facility location, submodular coverage, auditability}

\section{Motivation and contribution}

Shared-micromobility parking-zone design seeks demand coverage under exclusions, retained sites,
spacing, and area-level allocations. Our collaboration with the City of Braunschweig established
requirements to revise those constraints and compare feasible alternatives over common demand and
candidates. Each edit requires reoptimization. The municipal quality
agreement permits riding and parking restrictions to be
revised at any time, including temporary no-return areas for major events
\cite{StadtBraunschweig_2021}. Full-set greedy takes tens of seconds per alternative at city scale;
CLIPPER targets seconds-level feedback while retaining explicit constraints, replay records, and
missed-gain auditing. Planning-support research treats such what-if exploration as an aid to
deliberation \cite{Klosterman_1997,Pelzer_2014}.

Maximal covering and monotone submodular optimization provide transparent siting models
\cite{Church_ReVelle_1974,Nemhauser_1978}. Lazy, stochastic, and thresholded greedy methods reduce
marginal-query cost \cite{Badanidiyuru_Vondrak_2014,Mirzasoleiman_2015,Krause_Golovin_2014}.
Lazy evaluation retains the full candidate universe and skips redundant gain calculations; CLIPPER
instead restricts the candidates evaluated in each round and checks every active constraint.
Stochastic and thresholded variants trade queries for approximation under their stated constraint
models. Dynamic and prediction-augmented methods instead maintain solutions across element-update sequences
\cite{Agarwal_Balkanski_2024}. The closest dynamic method assumes a cardinality constraint, whereas
our setting combines retained sites, area-level caps, conflict classes, and network spacing. Prior
micromobility work has studied geofenced parking and strategic facility placement
\cite{Zhao_Ong_2021,Cai_2023_Geofencing,Teusch_2025}. CLIPPER (Constraint-exact Low-latency
Iterative Planning with Pooled Evaluation and Replay) provides a system-level execution contract.
This contract requires every run to construct a bounded pool, compute exact current gains, check
feasibility, and record the result for replay. Candidate generation remains replaceable; matched
full-set controls and separate screening and audit paths make the restriction inspectable. Unlike
methods that update a solution across edits, CLIPPER recomputes each recorded state.

\section{CLIPPER}

\begin{figure}[t]
  \centering
  \includegraphics[width=\columnwidth]{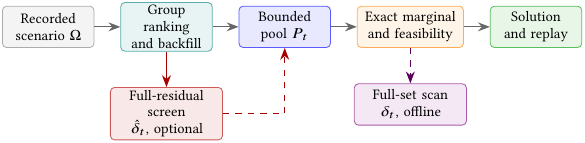}
  \caption{CLIPPER execution contract. \(\Omega\) is the recorded scenario; \(P_t\) is the pool at
  round \(t\); \(\hat\delta_t\) bounds the exact omitted gain \(\delta_t\). Solid arrows show timed
  execution; screening and full-set audits are excluded from rollout time.}
  \Description{A pipeline connects a recorded scenario, deterministic group ranking, a bounded
  candidate pool, exact feasible greedy selection, and replay output. A branch from the complete
  group ranking shows an optional full-residual screen; another shows an offline full-set audit.}
  \label{fig:short-pipeline}
\end{figure}

\paragraph{Planning model.}
Let \(V\) be the candidate set and \(\mathcal D\) the demand points with nonnegative weights \(w_d\).
Each \(e\in V\) covers \(C_e\subseteq\mathcal D\). For \(S\subseteq V\), the monotone submodular
objective is
\begin{equation}
  f(S)=\sum_{d\in\mathcal D}w_d\,
  \mathbf 1\!\left[d\in\bigcup_{e\in S}C_e\right].
  \label{eq:coverage}
\end{equation}
where \(\mathbf 1[\cdot]\) is the indicator; reported coverage is \(f(S)\) as a percentage of total
demand weight. A recorded scenario \(\Omega\) specifies active candidates \(V_\Omega\subseteq V\), locks
\(L(\Omega)\subseteq V_\Omega\), and a feasible family \(\mathcal I(\Omega)\subseteq2^{V_\Omega}\).
Feasible sets contain every lock,
exclude all unlocked candidates inside exclusion zones, obey the global budget and accounting-group
caps, take at most one site per conflict class, and satisfy minimum network spacing.
Locks override exclusions; evaluated scenarios are lock-feasible. CLIPPER seeks an
\(S\in\mathcal I(\Omega)\) maximizing \(f(S)\). The control is full-set greedy under the same policy;
reported gaps compare two greedy trajectories.
\Cref{fig:short-pipeline} gives the execution contract.

\paragraph{Restricted selection with exact feasibility checks.}
Proposal groups partition the candidates. Within each group,
CLIPPER ranks candidates once by singleton coverage \(f(\{e\})\), with stable tie-breaking. At
round \(t\geq1\), the incumbent is \(S_{t-1}\), initialized by \(S_0=L(\Omega)\). CLIPPER filters
invalid or selected candidates, backfills a bounded pool \(P_t\), and chooses
\begin{equation}
 e_t\in
 \underset{\substack{e\in P_t\\S_{t-1}\cup\{e\}\in\mathcal I(\Omega)}}{\arg\max}
 \Delta(e\mid S_{t-1}),\quad
 \Delta(e\mid S)=f(S\cup\{e\})-f(S).
 \label{eq:selection}
\end{equation}
and sets \(S_t=S_{t-1}\cup\{e_t\}\).
Thus shortlisting determines which candidates receive exact marginal evaluations, while the
selector checks every active hard constraint before insertion. This guarantees feasibility, not
optimality. Proposal groups map to accounting groups, whose caps constrain selection. CLIPPER-F uses
a fixed per-group width \(K\) under a policy that divides the residual facility budget equally among
accounting groups.
CLIPPER-A distributes one total candidate budget, favoring groups with more feasible candidates and
larger maximum singleton scores. Its coverage-prioritized policy relaxes candidate-count-weighted
accounting caps by \(\lambda=1.5\) without changing the global facility budget. We compare CLIPPER-A
only with a full-set control under those same caps.

Because \(S_0=L(\Omega)\in\mathcal I(\Omega)\) and every accepted update is checked,
\(S_t\in\mathcal I(\Omega)\) follows by induction. In the evaluated instances, K-means constructs
the candidate groups; the implementation filters and deduplicates proposals within each group before
selection. The singleton order is computed once. At every round, CLIPPER refilters candidates,
backfills the pool, and recomputes exact current marginals to account for demand already covered.
Recorded inputs and deterministic tie-breaking make the execution replayable.

Let \(W_t=|P_t|\) after filtering and deduplication, \(T\) be the number of insertions, and
\(c=\max_{e\in V_\Omega}|C_e|\). Exact marginal work is \(O(c\sum_{t=1}^{T} W_t)\), versus
\(O(cT|V_\Omega|)\) for full-set greedy or an exact audit along the returned trajectory. Both
expressions describe gain calculations rather than measured end-to-end wall time.

\paragraph{Online screening versus offline audit.}
Along the returned trajectory, let \(m_t=\Delta(e_t\mid S_{t-1})\), and let \(m_t^*\) be the best
feasible exact marginal over the complete scenario-filtered candidate set at the same incumbent.
Then
\begin{equation}
  \delta_t=m_t^*-m_t\geq0
  \label{eq:audit}
\end{equation}
Here, \(\delta_t\) is the exact gain omitted by the pool at round \(t\). Computing \(m_t^*\) requires
full-set exact-marginal scans
and is in the same cost class as full-set greedy; we use it offline or at high-stakes checkpoints
and exclude it from the reported rollout times. This round-level audit compares current gains along
the returned trajectory; complete executions provide the terminal coverage comparison.

For nonnegative weighted coverage, \(\Delta(e\mid S)\leq f(\{e\})\). CLIPPER can therefore reuse
cached singleton scores and scenario masks to compute a conservative residual screening bound
\[
  u_t=\max_{\substack{e\in V_\Omega\setminus S_{t-1}\\
   S_{t-1}\cup\{e\}\in\mathcal I(\Omega)}} f(\{e\}),\qquad
  \hat\delta_t=\max\{0,u_t-m_t\}.
\]
With \(u_t=0\) for an empty feasible residual set,
\(0\leq\delta_t\leq\hat\delta_t\). This avoids residual exact-marginal
recomputation but can still require filtering all residual candidates. CLIPPER uses this round-level
bound to trigger pool expansion or an offline audit.
If a pool has no positive feasible marginal before the budget is filled, expansion or a full scan
is required to distinguish true termination from shortlist truncation.

\paragraph{Replay contract.}
The replay record contains scenario and policy settings, aggregate counters, termination status, and
a selected-set fingerprint; configured audits add missed-gain values. Identical inputs and fixed
ranking rules, including tie-breaking, reproduce the fingerprint and aggregates.

\section{Evaluation}

\begin{figure*}[t]
  \centering
  \begin{minipage}[t]{0.24\textwidth}
    \centering
    \includegraphics[width=\linewidth]{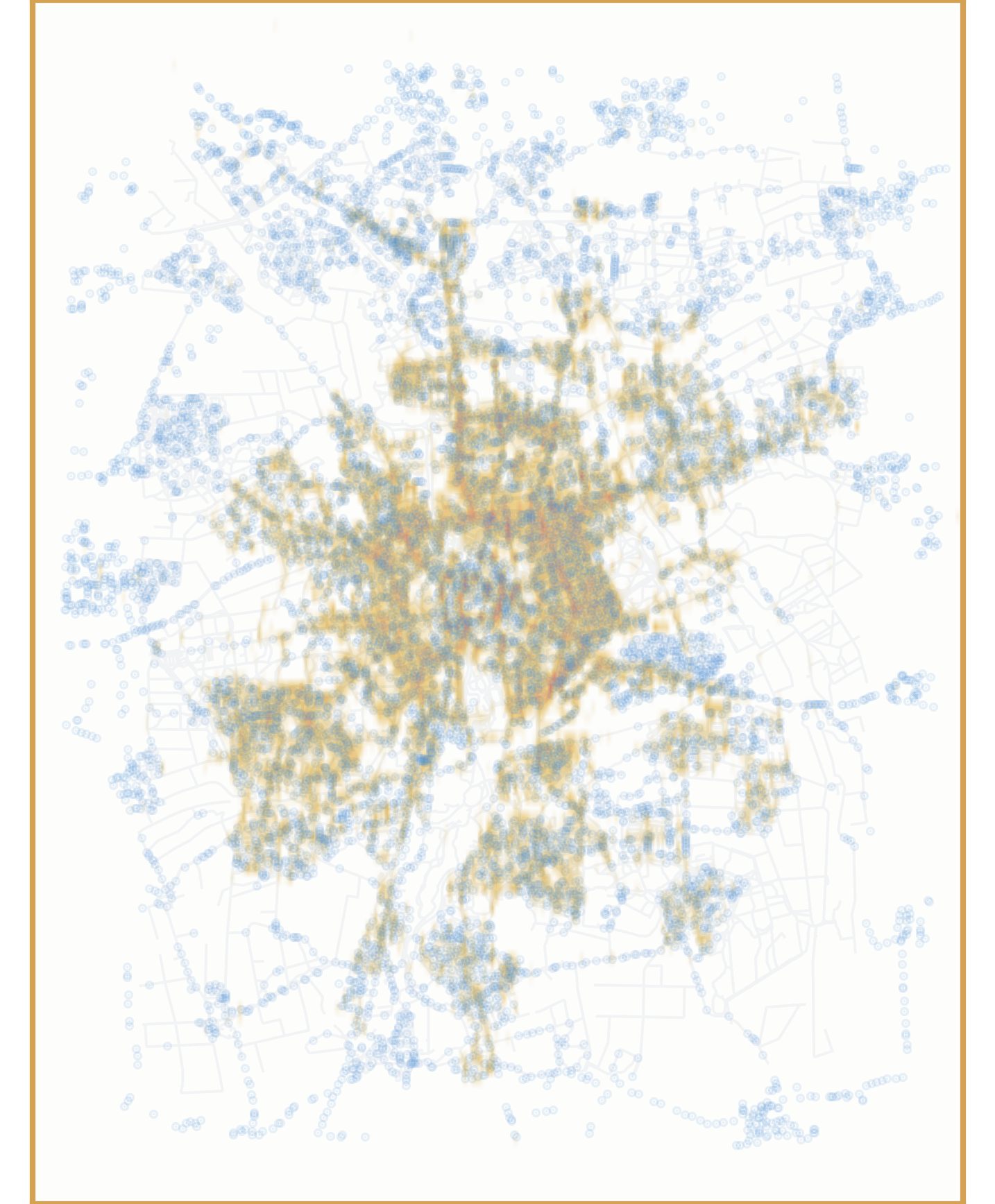}\\
    \scriptsize (a) Demand/support
  \end{minipage}\hfill
  \begin{minipage}[t]{0.24\textwidth}
    \centering
    \includegraphics[width=\linewidth]{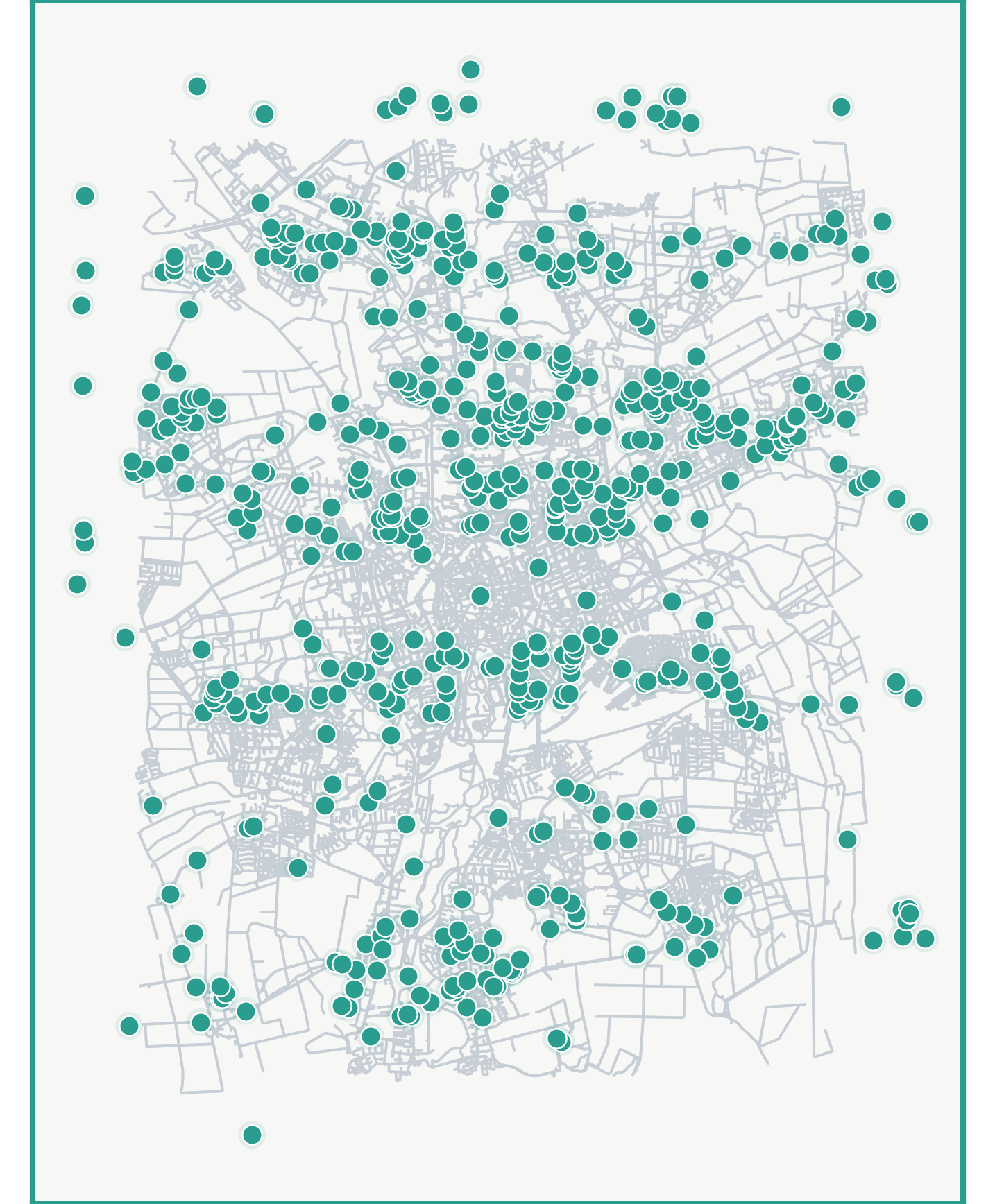}\\
    \scriptsize (b) Baseline \(E_0\)
  \end{minipage}\hfill
  \begin{minipage}[t]{0.24\textwidth}
    \centering
    \includegraphics[width=\linewidth]{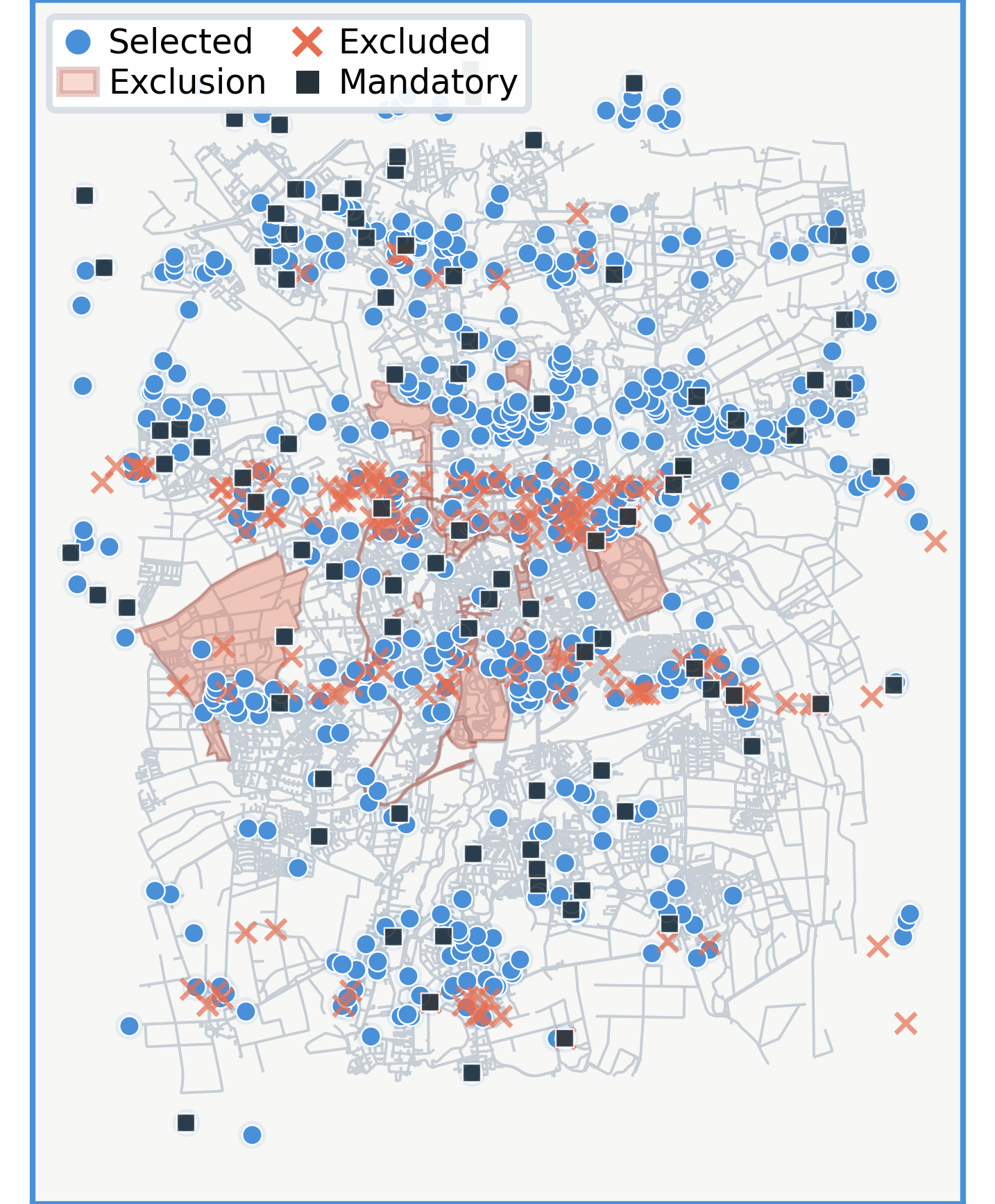}\\
    \scriptsize (c) Edited \(E_5\)
  \end{minipage}\hfill
  \begin{minipage}[t]{0.24\textwidth}
    \centering
    \includegraphics[width=\linewidth]{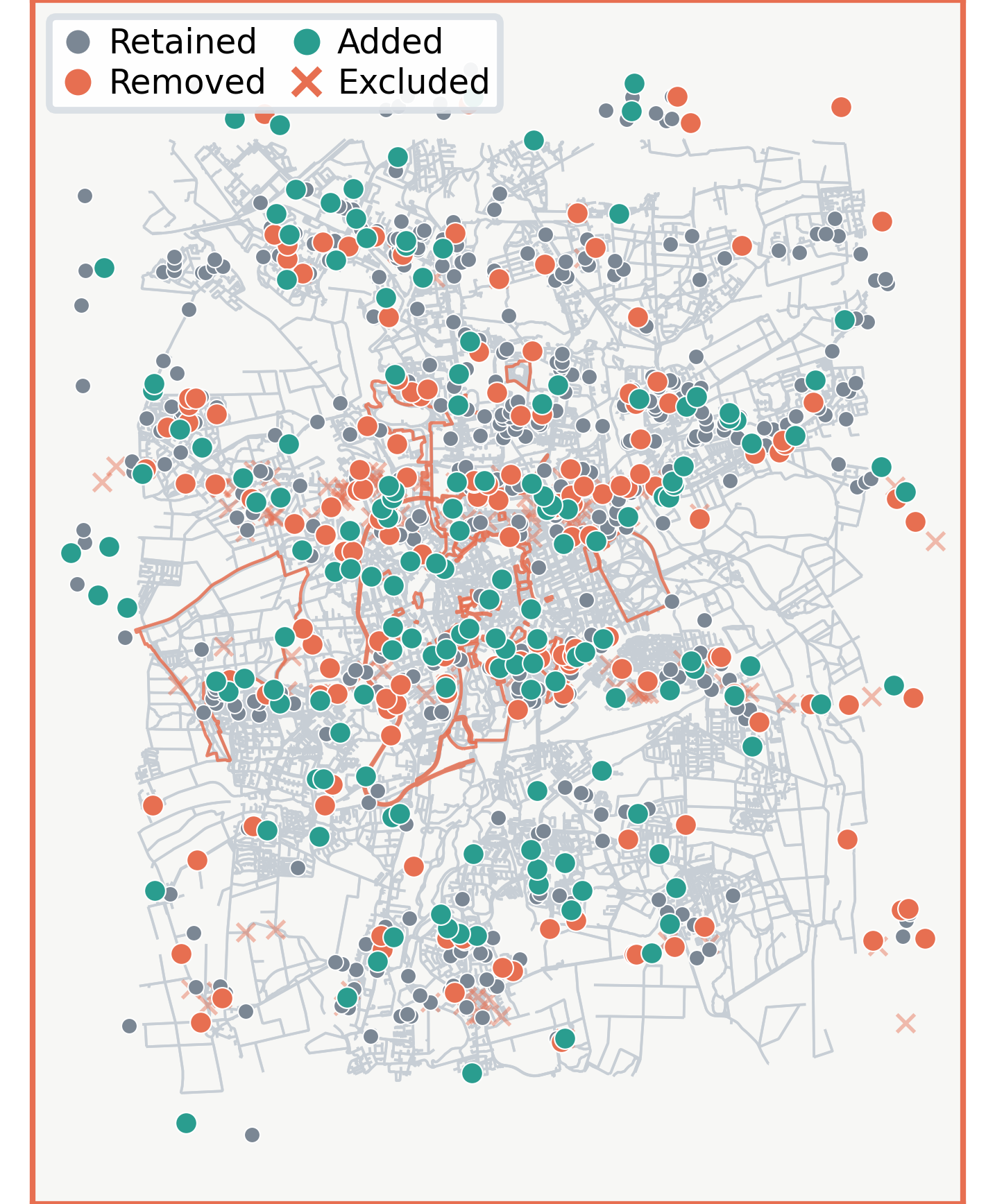}\\
    \scriptsize (d) \(E_0\)-to-\(E_5\) change
  \end{minipage}
  \caption{Braunschweig CLIPPER-F diagnostic replay using a \qty{25}{m} candidate-grid resolution.
  At \(E_5\), \num{162} candidates fall inside exclusions, \num{90} sites are mandatory, and
  \qty{25}{m} spacing applies; the plan retains
  \num{443} baseline sites and selects \num{157} replacements. Panel (d) shows retained, removed,
  and added sites. The benchmark edits are constructed rather than recorded municipal actions.}
  \Description{Four maps show demand and candidate support, the baseline selected plan, the
  edited plan with exclusions, locks, and spacing, and retained, removed, and added sites.}
  \label{fig:edit-example}
\end{figure*}

Recorded scenarios \(\Omega_t\) are denoted \(E_t\) for \(t=0,\ldots,10\). \Cref{fig:edit-example}
uses a \qty{25}{m} candidate grid in Braunschweig for legibility; runs use \qty{10}{m} there and
\qty{25}{m} elsewhere.

\begin{table*}[t]
  \centering
  \footnotesize
  \setlength{\tabcolsep}{4.0pt}
  \renewcommand{\arraystretch}{1.04}
  \begin{tabular*}{\textwidth}{@{\extracolsep{\fill}}lllrrrr@{}}
    \toprule
    Mode & Policy & City & Coverage: CLIPPER/control (\%) & Gap &
      Rollout: CLIPPER/control (s) & Speedup \\
    \midrule
    CLIPPER-F (\(K=1024\)) & balanced & BS  & 90.1/90.4 & 0.245 & 1.762/24.029 & \(13.6\times\) \\
                    &          & MUC & 69.7/69.7 & 0.003 & 1.493/22.943 & \(15.4\times\) \\
                    &          & BER & 59.5/59.5 & 0.001 & 1.825/52.685 & \(28.9\times\) \\
    \midrule
    CLIPPER-A (\num{8192} slots) & relaxed  & BS  & 95.0/96.8 & 1.82 & 3.219/28.343 & \(8.8\times\) \\
                    &          & MUC & 81.1/81.2 & 0.12 & 3.360/22.931 & \(6.8\times\) \\
                    &          & BER & 69.3/69.5 & 0.27 & 4.415/48.826 & \(11.1\times\) \\
    \bottomrule
  \end{tabular*}
  \caption{Full-data means over \(E_0,\ldots,E_{10}\). Paired cells give CLIPPER/control; gap is
  control minus CLIPPER coverage in percentage points, and speedup is control/CLIPPER rollout time.
  BS/MUC/BER denote Braunschweig/Munich/Berlin. Coverage is rounded to one decimal; gaps use
  unrounded means. CLIPPER-F and CLIPPER-A use different cap policies.}
  \Description{Six rows compare the two CLIPPER modes with full-set greedy under matching policies
  in Braunschweig, Munich, and Berlin. Columns report coverage, coverage gaps, rollout times, and
  speedups.}
  \label{tab:short-results}
\end{table*}

\paragraph{Data construction and edits.}
The benchmark retains every valid record in three non-public operator trip feeds available to the
project for internal shared-mobility research. Processed inputs remove provider identifiers. After
validation, the feeds contain \num{243649} Braunschweig trips
(2024-01-01--2024-04-30), \num{4958963} Munich trips (2023-12-18--2024-08-22), and
\num{12306947} Berlin trips over the latter interval. We snap each valid origin and destination to
its nearest feasible support point and assign one unit of demand to each snapped endpoint; we
aggregate endpoint demand over time for the spatial objective. Candidate grids use OpenStreetMap (OSM)-derived
sidewalk, footway, and public-parking support \cite{Haklay_Weber_2008} after restricted-area
filtering, at candidate-grid resolutions of \qty{10}{m} in Braunschweig and \qty{25}{m} in Munich
and Berlin. We initialize \(16\) clusters with K-means (seed \(42\)), then split or merge them until
each proposal group contains \(\num{1750}\text{--}\num{2750}\) candidates. The resulting instances have
\(29\) groups over \num{60495} candidates in Braunschweig; \(16\) over \num{32907} in Munich; and
\(36\) over \num{68922} in Berlin. These
groups define quota accounting and are not administrative districts. Coverage and facility spacing use
shortest paths on per-city OSM walk graphs, with a \qty{200}{m} coverage radius and
facility budgets of \(600\) (BS), \(1200\) (MUC), and \(1600\) (BER). At \(K=1024\), the maximum
proposal counts are \num{29696} (BS), \num{16384} (MUC), and \num{36864} (BER) per round; CLIPPER-A uses
\(8192\) total slots. All runs use the complete data.

We rerun every method from scratch on all states of the constructed stress chain.
Relative to \(E_0\), the chain increases the
share of high-contribution baseline sites receiving \qty{75}{m} exclusions from \(0\) to
\(17.5\%\), and the locked baseline share from \(0\) to \(35\%\).

\begin{samepage}
\paragraph{Edit construction.}
For each state, synthetic locks form a growing prefix of baseline sites ordered by contribution and
spatial dispersion. \Cref{tab:short-state-matrix}
records the exact schedule: \(E_5,\ldots,E_{10}\) ramp network spacing
from \qty{25}{m} to \qty{60}{m}, and \(E_8,\ldots,E_{10}\) add hotspot exclusions.
\end{samepage}

\paragraph{Results.}
On an Intel Core Ultra 9 285K, rollout time starts after scenario initialization and covers pool
construction, exact gains, feasibility checks, and selection. It excludes the preliminary
proposal-shape check, preprocessing, optional screening and audits, and serialization. \Cref{tab:short-results}
reports means across all 11 states. At the operating point \(K=1024\), CLIPPER-F stays within
\(0.245\) percentage points of the full-set control and reduces rollout time from
\(22.9\text{--}52.7\) to \(1.49\text{--}1.83\) seconds. CLIPPER-A with \num{8192} slots uses
\(9\text{--}15\%\) of the relaxed-cap control's time, with gaps of \(1.82\) (BS), \(0.12\) (MUC),
and \(0.27\) (BER) percentage points.
The CLIPPER-F and CLIPPER-A rows answer different policy questions because they use balanced and
coverage-prioritized caps, respectively. Summed over all 11 states of one city, the tabulated means
correspond to \(16.4\text{--}20.1\) seconds with CLIPPER-F versus \(4.2\text{--}9.7\) minutes with
the full-set control. This difference quantifies savings across repeated scenario comparisons.

\begin{table*}[t]
  \centering
  \scriptsize
  \setlength{\tabcolsep}{2.7pt}
  \begin{tabular*}{\textwidth}{@{\extracolsep{\fill}}lrrrrrrrrrr@{}}
    \toprule
    & \multicolumn{4}{c}{Constructed edit schedule}
    & \multicolumn{2}{c}{Braunschweig}
    & \multicolumn{2}{c}{Munich}
    & \multicolumn{2}{c}{Berlin} \\
    \cmidrule(lr){2-5}\cmidrule(lr){6-7}\cmidrule(lr){8-9}\cmidrule(l){10-11}
    State & Core (\%) & Lock (\%) & Hot & Space (m)
      & Gap & Speedup & Gap & Speedup & Gap & Speedup \\
    \midrule
    \(E_0\)  & 0.0  & 0.0  & 0  & 0  & 0.377  & 12.8 & -0.010 & 15.7 & 0.002  & 28.7 \\
    \(E_1\)  & 2.5  & 0.0  & 0  & 0  & 0.377  & 40.7 & -0.010 & 47.9 & 0.002  & 81.4 \\
    \(E_2\)  & 5.0  & 0.0  & 0  & 0  & 0.377  & 39.5 & 0.016  & 49.1 & 0.002  & 84.3 \\
    \(E_3\)  & 5.0  & 10.0 & 0  & 0  & 0.319  & 40.7 & 0.004  & 46.5 & 0.002  & 72.5 \\
    \(E_4\)  & 7.5  & 10.0 & 0  & 0  & 0.319  & 38.4 & 0.004  & 44.7 & 0.002  & 72.7 \\
    \(E_5\)  & 7.5  & 15.0 & 0  & 25 & 0.312  & 10.4 & 0.004  & 11.3 & 0.000  & 21.2 \\
    \(E_6\)  & 10.0 & 20.0 & 0  & 35 & 0.266  & 9.2  & 0.004  & 10.0 & 0.000  & 19.9 \\
    \(E_7\)  & 12.5 & 20.0 & 0  & 45 & 0.243  & 9.4  & 0.004  & 10.4 & 0.000  & 19.3 \\
    \(E_8\)  & 12.5 & 25.0 & 10 & 50 & 0.079  & 8.4  & 0.004  & 9.3  & 0.000  & 18.0 \\
    \(E_9\)  & 15.0 & 30.0 & 15 & 55 & -0.038 & 8.0  & 0.004  & 8.7  & -0.001 & 17.0 \\
    \(E_{10}\) & 17.5 & 35.0 & 20 & 60 & 0.064  & 7.4  & 0.004  & 8.4  & -0.001 & 15.7 \\
    \bottomrule
  \end{tabular*}
  \caption{Complete chain and CLIPPER-F performance. Core is the baseline share assigned
  \qty{75}{m} exclusions; Lock is the mandatory share; Hot counts \qty{150}{m} exclusion halos; Space
  is minimum network spacing (zero disables it). Gap is control minus CLIPPER coverage in percentage
  points; speedup is control/CLIPPER rollout time. Each state is recomputed.}
  \Description{Eleven rows list the exclusion, mandatory-site, hotspot, and spacing settings of
  states E-zero through E-ten, followed by the CLIPPER-F coverage gap and speedup in each city.}
  \label{tab:short-state-matrix}
\end{table*}

\paragraph{Chain sensitivity and evidence scope.}
\Cref{tab:short-state-matrix} reports all \(33\) paired state--city outcomes so edit sensitivity
remains visible. For CLIPPER-F, per-state gaps to the control range from \(-0.038\) to \(0.377\)
percentage points in Braunschweig, from \(-0.010\) to \(0.016\) in Munich, and from \(-0.001\) to
\(0.002\) in Berlin. Per-city speedup ranges are \(7.4\text{--}40.7\times\) (BS), \(8.4\text{--}49.1\times\) (MUC), and
\(15.7\text{--}84.3\times\) (BER). Speedups are largest before network-spacing constraints activate at
\(E_5\), but remain at least \(7.4\times\) after spacing and hotspot edits are combined. Negative gaps
arise when restricted and full-set greedy follow different trajectories. The cross-city contrast
supports city-specific calibration of the per-group width \(K\) rather than a universal setting.

\paragraph{Determinism and audit scope.}
For CLIPPER-F with \(K=1024\), three full-data reruns of every state and city (\(99\) runs) produced
identical coverage, step and gain counts, termination reasons, and terminal flags within each state.
Every run exhausted all feasible positive gains; no run stopped because its pool was too narrow.

\section{Limitations and scope}

\paragraph{Limitations.}
Collaboration with the City of Braunschweig informed the requirements; the controlled
\(E_0,\ldots,E_{10}\) chains evaluate optimizer behavior rather than user interaction or deployment
and do not measure edit frequency, response thresholds, or planner utility. The coverage model omits
congestion, compliance, equity, rebalancing, and curb capacity. We evaluate one operating point per
mode with a static ranking; refreshed rankings and transfer remain open. The trip feeds are incomplete
and cannot be redistributed. Public reconstruction also requires processed candidates, scenario
records, and fixed OSM snapshots.

\paragraph{Operational interpretation.}
CLIPPER separates scenario design from optimization: planners define a policy state, the optimizer
returns a feasible plan over common data, and replay records preserve its context. Fixed demand,
candidates, and network data make output differences traceable to policy edits. A comparison can
show coverage, constraints, selected-site changes, and termination status. Before interactive use,
a planning team can test several values of \(K\) on stored states against the same-policy full-set
control and retain the smallest width meeting its coverage-gap target. The chosen \(K\), tie rules,
and input versions define the replayable configuration. A planner study should measure time to an
accepted scenario, alternatives inspected, corrections, and record utility.

\section{Conclusion}
CLIPPER makes repeated planning runs inspectable. Bounded pools reduce each alternative from
tens of seconds to seconds while exact checks enforce every encoded constraint. Each pool policy is
judged against full-set greedy under the same policy. Matching the cap policy separates candidate
restriction from differences in allocation rules. Replay records preserve scenario context, and
offline scans reveal omitted gains. The execution contract allows proposal rules to change without
altering feasibility or comparison semantics. The practical unit of comparison is therefore not a
score alone, but a versioned policy state, its feasible plan, and the record needed to rerun it.

\begin{acks}
We thank our partners at the City of Braunschweig for contributing operational requirements.
Generative AI tools were used only for editing and language polishing. The authors take full
responsibility for the scientific content, experiments, analyses, and conclusions.
\end{acks}

\bibliographystyle{unsrtnat}
\bibliography{references}

\end{document}